\documentclass[10pt,journal,compsoc]{IEEEtran}
\ifCLASSOPTIONcompsoc
  \usepackage[nocompress]{cite}
\else
  \usepackage{cite}
\fi
\usepackage{algorithm}
\usepackage{algorithmic}
\usepackage{times}
\usepackage{epsfig}
\usepackage{graphicx}
\usepackage{amsmath}
\usepackage{amssymb}
\usepackage{enumerate}
\usepackage{subfig}
\usepackage{booktabs}
\usepackage{dsfont}

\ifCLASSINFOpdf
\else
\fi
\begin{document}
%
\title{Perceptual Visual Interactive Learning}
%
%
%
%


\author{Shenglan~Liu,~\IEEEmembership{Member,~IEEE,}
		Xiang~Liu,
		Yang~Liu,
		Lin~Feng,
		Hong~Qiao,~\IEEEmembership{Senior~Member,~IEEE}
        Jian~Zhou,
		Yang~Wang
	
\IEEEcompsocitemizethanks{\IEEEcompsocthanksitem Shenglan Liu,
	Xiang Liu, Yang Liu, Lin Feng and Yang Wang are with Faculty of Electronic Information and Electrical Engineering, Dalian University of Technology, Dalian, Liaoning, 116024 China  e-mail: (\{liusl, dlut\_liuyang, zhoujian, wangyang521\}@mail.dlut.edu.cn, fenglin@dlut.edu.cn, xliudut@gmail.com). Lin Feng is the Corresponding author.\IEEEcompsocthanksitem H. Qiao is with the State Key Laboratory of Management and Control for Complex Systems, Institute of Automation, Chinese Academy of Sciences, Beijing 100190 China  e-mail: (hong.qiao@ia.ac.cn).
}

}

%
%

\markboth{Journal of \LaTeX\ Class Files,~Vol.~14, No.~8, August~2015}%
{Shell \MakeLowercase{\textit{et al.}}: Bare Demo of IEEEtran.cls for Computer Society Journals}
%



\IEEEtitleabstractindextext{
\begin{abstract}
	
Supervised learning methods are widely used in machine learning.
However, the lack of labels in existing data limits the application of these technologies.
Visual interactive learning (VIL) compared with computers can avoid semantic gap, and solve the labeling problem of small label quantity (SLQ) samples  in a groundbreaking way.
In order to fully understand the importance of VIL to the interaction process, we re-summarize the interactive learning related algorithms (e.g. clustering, classification, retrieval etc.) from the perspective of VIL.
Note that, perception and cognition are two main visual processes of VIL.
On this basis, we propose a perceptual visual interactive learning (PVIL) framework, which adopts gestalt principle to design interaction strategy and multi-dimensionality reduction (MDR) to optimize the process of visualization.
The advantage of PVIL framework is that it combines computer's sensitivity of detailed features and human's overall understanding of global tasks.
Experimental results validate that the framework is superior to traditional computer labeling methods (such as label propagation) in both accuracy and efficiency, which achieves significant classification results on dense distribution and sparse classes dataset.

\end{abstract}

}

\maketitle

\IEEEdisplaynontitleabstractindextext

%
\IEEEpeerreviewmaketitle

\IEEEraisesectionheading{\section{Introduction}\label{sec:introduction}}

%
%
%
%

\IEEEPARstart{H}uman vision and perception play an important role in data preprocessing, which has been illustrated in many previous works\cite{Oliveira2003From}\cite{Keim2002Information}\cite{Keim2006Challenges}.
Visual interactive learning methods, transmitting human's perceptual information of data to computer, are different from traditional machine learning methods\cite{Demir2011Batch}.
Such as the semantic gap of image retrieval \cite{Hare2006Mind}, the semantic segmentation of images \cite{Long2014Fully} and so on.
In image retrieval, relevant feedback labels related (or unrelated) images of human cognitive according to the original ranking list of query, which constructs a new ranking model to rerank the images to avoid the image semantic gap as much as possible\cite{Salton1988Improving}.
In image segmentation, Li et al. utilize human visual perception and cognition of images to achieve semi-supervised segmentation through interaction between humans and images \cite{Li2004Lazy}.
Besides, human-computer interaction related studies have improved the effect of image processing through eye movement \cite{Fisher2017Eye} and region of interest (ROI)\cite{Brett2002Region}. In general, the key point of visual interactive learning is extracting important information (e.g. position, contour) from data or images using visualization.
Perception and cognition are two main visual processes of humans receiving information.
Depending on existing technology (e.g. image segmentation, image retrieval), we re-summarize visual interactive learning from the aspects of perception and perception with cognition.
Visual interactive learning is defined as a method of machine learning which interacts data using human's perception and cognition.

It is illustrated that visual interactive labeling is a key issue in visual interactive learning, which is always used in machine learning methods such as classfication\cite{Demir2011Batch}, retrieval\cite{Moumtzidou2016VERGE}, and clustering\cite{Awasthi2014Local}.
In recent years, many related machine learning methods have been proposed. However, the SLQ problems limit the development of these technologies, especially in the classification method where a large number of labels are needed, such as deep learning methods \cite{Farabet2013Learning}\cite{Lecun2015Deep}.
Therefore, labeling is an important issue that is difficult to solve in SLQ machine learning. The traditional classification framework for SLQ is generally classified by support vector machine (SVM)\cite{Chapelle1999Support} after expert labels or active learning \cite{Bakas2017Advancing}\cite{Brinker2006On}.
However, this approach is expensive in labeling and calculating, so we hope to develop an efficient interactive learning method to solve the SLQ problem.

Classification is one of the major applications of data labeling, which can generally be fallen into non-metric models and metric-related models.
Non-metric models can be regarded as perceptually unrelated models, which are difficult to understand with human perception, i.e. iterative dichotomiser 3 (ID3) \cite{Umanol1994Fuzzy}, classification and regression tree (CART) \cite{Sahin2011Detecting} and other decision tree methods.
The other is metric-related classification approaches, which are easy to visualize and conform to the human perception system.
One typical classification model is k-nearest-neighbor (KNN) classifier \cite{Zhang2006SVM}.
The classical KNN model is based on the $L_2$-norm metric.
To improve the performance of KNN, some extended versions (e.g. $L_1$-norm based KNN, extended nearest neighbor (ENN) and cosine metric based KNN \cite{Liu2017Perceptual}\cite{Feng2013Maximal}) are proposed according to KNN framework. However, lazy learning methods (including KNN and the extended versions) strong relate to the distribution and the number of training samples (KNN -based methods are always difficult to apply to large-scale data sets and are quite time consuming.).

On this basis, many model-based classification methods such as SVM \cite{Chapelle1999Support}, random vector functional link (RVFL) \cite{Husmeier1999Random} and other discriminant models are proposed, which only needs to learn a linear or nonlinear function to achieve fast classification.
Since then, with the rapid development of digitalization, neural network classifiers under big data, such as the deep learning (DL) methods, have been widely used and have achieved outstanding results on ImageNet \cite{NIPS2012_4824}\cite{Simonyan2014Very}\cite{He2016Deep}.
However, labeling is the bottleneck of the DL methods in big data. Therefore, data labeling has become an important issue that needs to be resolved in the past two years.
Traditional data labeling is a machine-to-machine learning method, such as label propagation method \cite{Zhu2005Semi}, utilizing the potential manifold structure and nonlinear metrics of the data to pass label to the unlabeled samples, can be considered as a classification method for SLQ samples.


In recent years, data labeling has achieved some remarkable results, especially in visualization labeling field.
The categories of labels include video, text, time series, and scatter\cite{Bernard2017Visual1}.
For example, Hoferlin used a specific interactive classifier for video visualization analysis \cite{Heidemann2013Inter}.
Heimer et al. reduced the workload of obtaining labels in text search through interactive training\cite{Heimerl2012Visual}.
Sarkar and Bernard et al. obtained relevant labels for semi-supervised interaction models for industrial data \cite{Sarkar2016Visual} and human motion time series data \cite{Bernard2017Visual2}, respectively.
Sedlmair et al. proposed an interactive model for measuring scatter-ability, correlation, and outliers\cite{Sedlmair2015Data}.
In addition, the types of labeling include classification tasks, relevant feedback, correlation coefficients, etc.
For example, relevant feedback in the field of image retrieval can effectively improve the accuracy of retrieval \cite{Salton1988Improving}\cite{josephrocchio}\cite{Rui1998Relevance}.
Bernard etc. proposed a method that characterizing the patient's health index by actively learning the doctor's feedback \cite{bernard2015visual}.
To enhance the performance of learning tasks, human-to-machine approaches are proposed by combining interactive learning with other machine learning methods.
For example, interactive labels are used to reflect similarities in metric learning, and to interpret the interrelationships between instances from user interaction features \cite{bernard2017visual}.
Bernard etc. pointed out that interactive labeling is superior to active learning given the condition the class distributions are separated well in dimension reduction\cite{Bernard2017Comparing}.

The existing interaction methods are main focus on expanding the amount of labels available, improving learning result and reducing the complexity of manual labeling \cite{Bernard2017Comparing}\cite{Bernard2017VIAL}, which  are instructive and excellent on designing interaction strategies based on existing technical characteristics.
However, most of these works are lack of considering human's visual processing advantages, and limit the further application of visual interaction.
At present, there are few interactive learning methods that combine computer process and human perception.
This paper points out two important issues that interactive learning should notice:
(1) human perception of data exists in a manifold, which may have significant impact on visualization and interaction (e.g. overlapping, bending manifold).
(2) human has a strong visual perception over data (especially large amount of data).

In response to the above issues, This paper focuses on the method of visual interactive labeling
based on human perception in SLQ datasets and proposes multi-feature selection and multi-dimension reduction approach, which makes full use of the discriminative features of data manifolds, wakens the influence of coupling on visualization, and achieves better labeling results.
Meanwhile, this paper introduces Gestalt principle to understand human perception.
Gestalt principles are derived from the process of human understanding of the world and have guiding significance for the design of interaction strategies.
Based on the understanding perception and data manifold, we propose a perceptual visual interactive learning framework.
The framework considers SLQ samples in semi-supervised learning, which significantly improve the efficiency and accuracy for classification.
The key contributions of this paper to interactive learning can be summarized as follows.

(1) This paper re-summarizes visual interactive learning (VIL) from the aspects of human visual perception and cognition, and makes a division of VIL, which has important significance for comprehensively understanding the visual-based interactive technology.

(2) This paper points out the reason for the overlapping visual data, and proposes a user-oriented visualization method for multi-feature selection and multi-dimension reduction.

(3) Compared with existing labeling methods (e.g. label propagation), human have a strong visual perception ability over little amount of labels or large amount of data. We point out and analysis the above point of view, and illustrate the PVIL framework proposed in this paper superior to the traditional labeling method in correctness and time consumption, and has a better classification effect on real-world dataset.

(4) PVIL framework adopts the Gestalt principle to explain several criteria for human perception, by which the design process of the interaction strategy is guided.
It provides a theoretical basis for visual interactive learning in human perception.


The rest of this paper is organized as follows. Section \ref{relatedintro} introduces the computer labeling and human perception.
Section 3 summarizes visual interactive learning and introduces our PVIL framework.
The experimental results and analysis of this paper are listed in section 4, which gives a comparison between the PVIL framework and computer labeling in both accuracy and efficiency aspects, and the semi-supervised classification experiment is also conducted. The last section is conclusion.

\section{Related Work}

\label{relatedintro}

In VIL community, computer labeling cooperates with human perception.
As a typical method of computer labeling, label propagation\cite{Zhu2005Semi} is briefly introduced in section 2.1.
Human perception principles for classification labeling are based on the Gestalt principles\cite{Pelli2009Grouping}\cite{Demany1977Rhythm}, which can be found in section 2.2.

\subsection{Label Propagation (LP)}

\label{related1}

Expert labeling, as a common solution to obtain labels, is complicated and time-intensive.
By contrast, the key idea of semi-supervised learning is to employ a large amount of unlabeled data and partially labeled data to structure a classifier that achieves relatively high accuracy with less human participants\cite{Chapelle1999Support}.
As a typical semi-supervised learning method, LP\cite{Zhu2005Semi} algorithm generate labels with the help of the graph weight matrix.

Giving a $n$ samples dataset $X=[x_1,\cdots,x_n]\in \mathds{R}^{D \times n}$, where $D$ indicates the dimensions of one sample. LP algorithm develops a fully connected graph by the labeled set $D_l \in \mathds{R}^{l}$, corresponding labeled samples $X_l \in \mathds{R}^{D \times l} $ (Eq. (\ref{eq1})) and the unlabeled set $X_u \in \mathds{R}^{D \times u}$ (Eq. (\ref{eq2})), respectively.

\begin{equation}
\begin{array}{l}
X_l = [x_0, \cdots, x_l]\\\\
D_l = [y_1, \cdots, y_l]
\end{array}
\label{eq1}
\end{equation}

\begin{equation}
X_u = [x_{l+1},\cdots,x_{n}]
\label{eq2}
\end{equation}

where $x_i \in \mathds{R}^D$, $i=1,2, \cdots, n,$  is a representation of the instance, $y_j \in \{c_1,c_2,\cdots,c_p\}$, $j=1,2,\cdots,l$ is the class label of the $j$-th sample ($p$ indicates the number of the classes), and $n=l+u$ (always $l \ll u$).

The weight matrix $W = \left\{ {W_{ij} } \right\} \in { \mathds{R}}^{n\times
n}$ describing weight of each pair of different samples $x_m$ and $x_j$ can be denoted by Eq. (\ref{eq122}) as follows.

\begin{equation}
w_{i,j} =
\begin{cases}
 exp(-\gamma||x_m - x_j||), x_m, x_j \in N_k(x_i)\\
0, else\\
\end{cases}
\label{eq122}
\end{equation}

where $\gamma$ is a hyper-parameter relating to the data distribution. Then, we compute the symmetrically normalize matrix of $W$, which denotes as $S = D^{{ - 1}
\mathord{\left/ {\vphantom {{ - 1} 2}} \right. \kern-\nulldelimiterspace}
2}WD^{{ - 1} \mathord{\left/ {\vphantom {{ - 1} 2}} \right.
\kern-\nulldelimiterspace} 2}$,
where $D$ is a diagonal matrix, and $D_{ii} = \sum\nolimits_{j = 1}^n {W_{ij}
} $. In  Eq.(\ref{eq122}), $||\cdot||$ can be $L$1/$L$2 or other norm.

We denote $Q=[q_1,\cdots,q_n] \in \mathds{R}^{n}$ as the original label vector which satisfies $q_i=y_i$, $i=1,2,\cdots,l$,  $q_j=0$, $j=l+1,l+2,\cdots,n$. $f^{t}$ indicates the $t$-th iteration of label vector, and $f^*$ is the final label result with labeled $q_j$, $j=l+1,l+2,\cdots,l+u$. The iteration equation of LP can be expressed by  Eq. (\ref{eq133}) as follows.
\begin{equation}
f^{t+1}=\lambda Sf^{t}+(1-\lambda)Q
\label{eq133}
\end{equation}

where $\lambda$ is a balance weight parameter of $Sf^{t}$ and $Q$. We can get $f^*$ by Eq. (\ref{eq133}) while $||f^{t+1}- f^{t}||<\varepsilon$, where $\varepsilon>0$ is a small real number to stop the iteration. LP, as a metric-based approach, shows the advantage of computational accuracy in computer labeling. However, in real word applications, most of computer labeling methods are not ideal, which can be attributed to the high complexity of time and space. Furthermore, the process of propagation is easily misled by neighbors.


\subsection{Gestalt Principle}

\label{related2}

Gestalt psychology arises from the research of perception (as famously noted by Koffka\cite{Koffka1935Principles}).
the process of perception could be understood as a series of principles, which have a closed relationship with human visual models, and then can be very widely used to assist the design of interactive models \cite{Zhu2005Semi}\cite{Pelli2009Grouping}\cite{Peterson2013The}\cite{Todorovic2008Gestalt}.
In this study, the theory of visual labeling involves three principles, which are proximity, closed and continuity respectively.

\begin{figure}[htbp]
	\centerline{\includegraphics[width=3.4in]{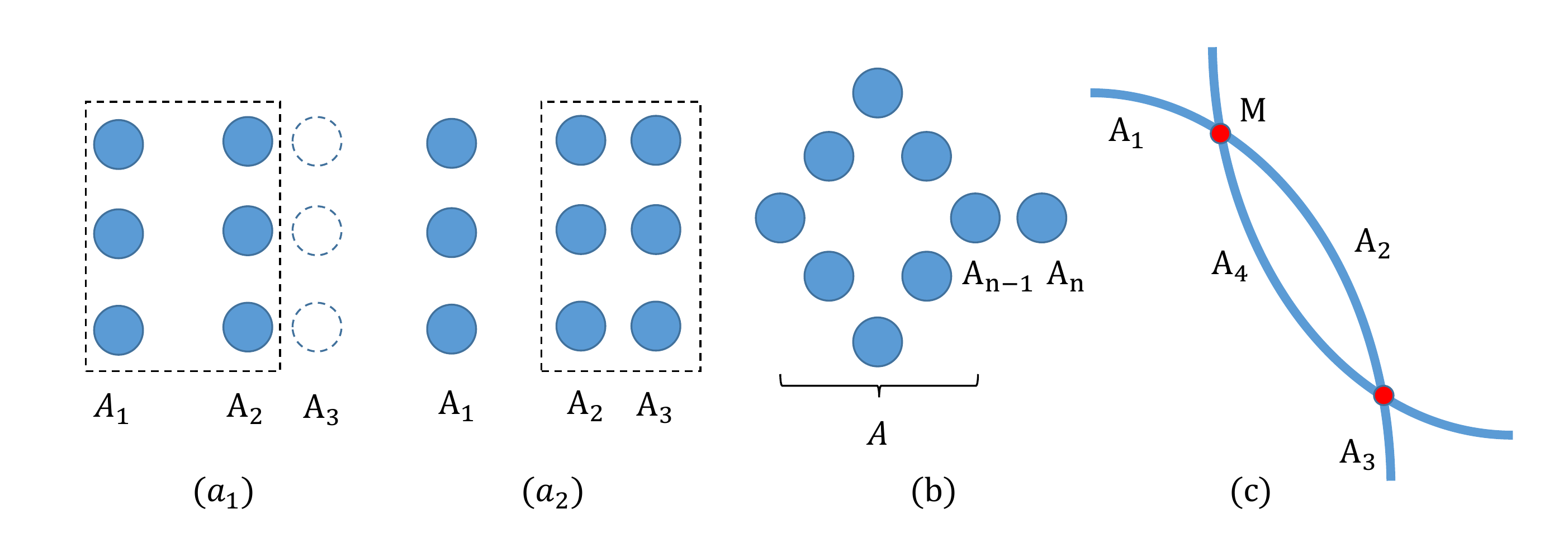}}
	\caption{Gestalt principle of human perception (a) proximity principle (b) closed principle (c) good continuity principle }
	\label{fig1}
\end{figure}

\subsubsection{Proximity Principle}
The proximity principle indicates that elements with proximate positions are tend to be classified into one category.
Assume that there are three independent elements, which are $A_1$, $A_2$ and $A_3$ respectively.
Based on this principle, the symbolized form can be written as Eq. (\ref{eq21}) .

\begin{equation}
\label{eq21}
[A_1, A_2], A_3 \to A_1, [A_2, A_3]
\end{equation}

As shown in Fig. \ref{fig1} ($a_1$) and Fig.\ref{fig1} ($a_2$), $A_1$ and $A_2$ could be classified into one group when $A_3$ does not exist.
Similarly when $A_3$ is added, $A_2$ and $A_3$ tend to be put together.
The different distance between elements results in different classification situation in Fig. \ref{fig1} ($a_1$) and Fig.\ref{fig1} ($a_2$).
It is worth noting that the distance measure plays a key role in evaluation indicator for various classification tasks.
The common unsupervised classifier KNN, for example, is focuses on the neighbor distance \cite{Zhang2006SVM}.
Also, the loss function of the linear classifier's gradient descent is based on the distance between the sample and the hyperplane \cite{Joachims1998Making}.

\subsubsection{Closed Principle}
The closed principle suggests that elements with integrated boundary are tend to classified into one category.
Suppose that the independent elements can be represented by $A_1$, $A_2, \dots, A_n$, where $A=\{A_1, A_2, \dots, A_{n-1}\}$.
Its symbolization can be expressed as Eq. (\ref{eq22}).

\begin{equation}
\label{eq22}
A_1,A_2,\dots,A_{n-1},A_n  \to \underbrace{[A_1,A_2,\dots,A_{n-1}]}_{\text{A}} , [A_n]
\end{equation}

As illustrated in Fig. \ref{fig1} (b), the closest element to the $A_{n-1}$  is $A_n$, while $A_{n-1}$ and the rest of the elements ($A_{1}, \cdots, A_{n-2}$) is classified into a same set, of course, $A_n$ is grouped independently.
Apart from $A_n$, the remaining elements can be gathered into a meaningful set $A$.
That is, set $A$ is consist with the priori combination hypothesis of a certain shape.
Moreover, in human version, set $A$ has a closed tendency.
In the process of perceiving the world, human vision tends to combine scattered individuals, in other words, to understand isolated elements from a macroscopic perspective \cite{Wandell1995Foundations}.
In visual analysis, the significant different between human and computer is that, the perception of humans towards real world can be viewed as a collection of elements.
As far as our knowledge is concerned, the advantage of human perception derived from its macroscopically discriminant ability, by sharp contrast, computer merely has ability to calculate relationships between certain isolated elements.

\begin{figure*}[htbp]
	\centerline{\includegraphics[width=5.8in]{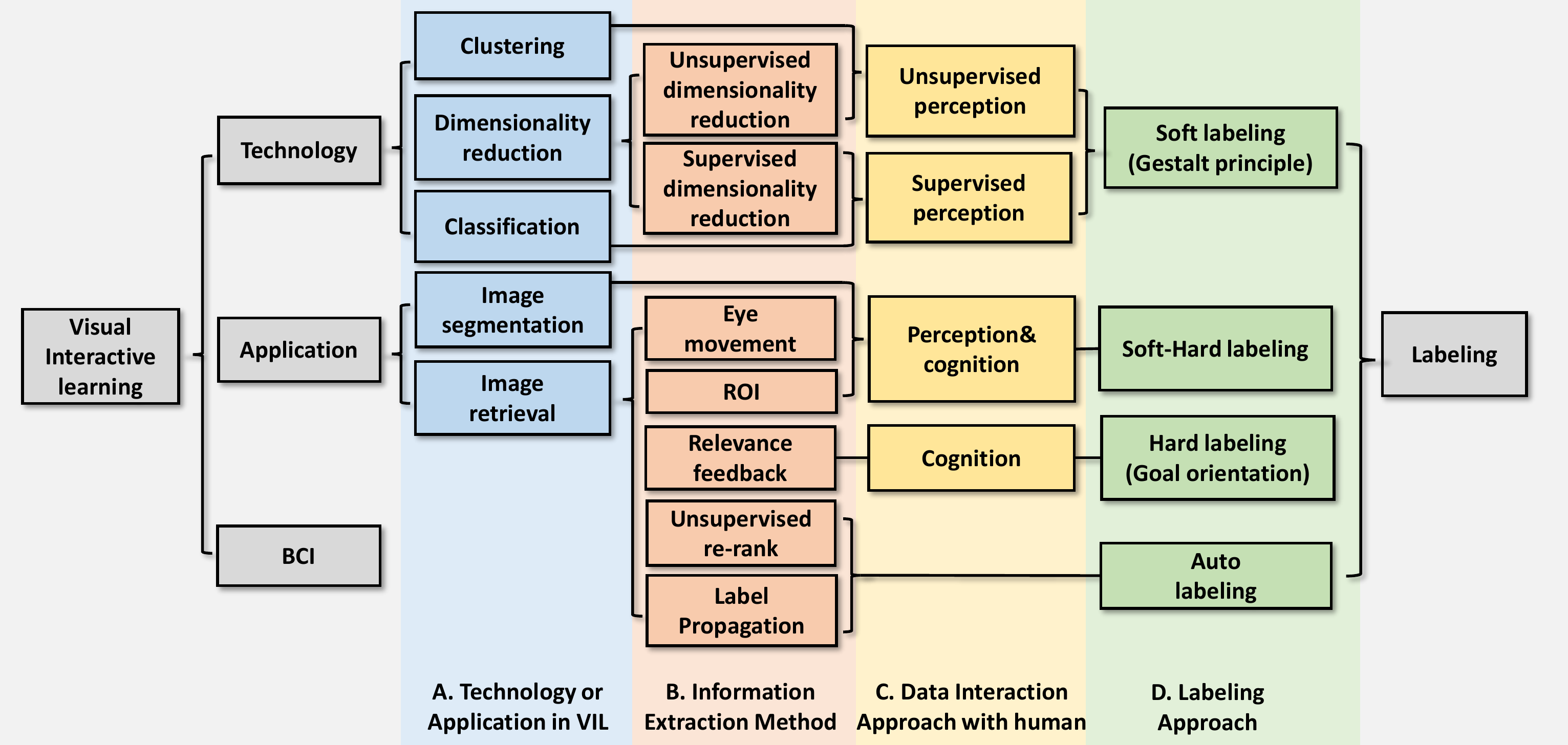}}
	\caption{The re-summarized visual interactive learning framework. From an interaction perspective, they can interact through perception interaction, cognition interaction, and combination of perceptual-cognitive interaction. We classify these labels into soft labeling, hard labeling, soft-hard labeling, and auto labeling.}
	\label{fig2}
\end{figure*}

\subsubsection{Good Continuity Principle}
Good continuity principle suggests that the elements with shared lines, curves or planes are tend to be seen as an integral.
In certain cases, An integrated element may be divided into several separated parts.
However, part of continuous information (e.g. curvature continuity information, dimensional continuity information, etc.) of the origin element is retained in the separated parts.
Good continuity principle indicates that human vision tend to recruit the integrated element by means of these continuous information.
Assuming there are several discrete elements, $A_1, A_2, A_3,$ and $A_4$ respectively, and $A$ is a visual-integrated element.
The symbolized form of good continuity principle can be denoted as Eq. (\ref{eq23}).

\begin{equation}
\label{eq23}
[A_1 + A_2 + A_3] = [A]
\end{equation}

As clearly revealed in Fig. \ref{fig1} (c), $A1$ and $A2$ are broken in $M$, and the retained continuous information mainly includes curvature continuity information.
Human vision tends to make judgments that treating a manifold with continuous curvature as a integral \cite{Wandell1995Foundations}.
The connection of $A_4$ and $A_1$ simply meet the first-order continuity in $M$, while the connection of $A_2$ and $A_1$ not only confirm to the first-order continuity but also meet the second-order continuity.
Therefore, $A_1$, $A_2$, and $A_3$ are assigned into a visual-integrated element $A$ ($A_2$ and $A_3$ are assigned in a similar way with $A_1$ and $A_2$).
The continuous information can not accurately perceived by computer in the process of calculation, which can be deemed as a disadvantage of computer in comparison with human perception.

\section{Visual Interactive Learning}

The key aspect of visual interactive learning is extracting useful information from data for machine learning by using visualization method.
In section 3.1, Visual interactive learning are re-summarized by perception and cognition of visual processes.
Section 3.2 introduces our multi-dimensionality reduction method which can solve data overlap problem.
The perceptual visual interactive learning framework is introduced in the last subsection.

\subsection{Visual Interactive Learning (VIL)}

The aim of VIL is to label samples of dataset from both computer's computing ability and human's visual understanding ability.
The three Gestalt principles, described in subsection 2.2, illustrate the superiority of human visual perception which motivates us to survey VIL on a new point of view.
We re-summarize the VIL according to distinction and relationship between computer and visual perception of human in Fig. \ref{fig2}.
According to the process of VIL, A, B, C and D are introduced respectively.

A mainly includes VIL related technologies (clustering, dimension reduction and classification) and applications (image segmentation and image retrieval).
In general, VIL interacts with human from the aspect of vision.
Of course, brain-signal-based BCI can also be deemed as a VIL method.
B is denoted as a information extraction method.
For the purpose of visualization, B maps samples to a certain visual space.
B could be seen as concrete realization of A, which includes eye movement, ROI and relevance feedback,
Depend on the method of B, C is indicated as a data interaction approach with human.
After data is mapped to an image in visual space, human could decide how to interact with the image after visualization through operability (perception) and requirement (cognition).
As an interaction approach, C mainly includes unsupervised perception, supervised perception, cognition, and combination of perception and cognition.
D is expressed as a Labeling approach, which can also be seen as a interactive result of C.
The information of perception and cognition from human are transmitted to computer by means of labels.
With the help of the labels, the model computed by computer is in line with human needs.
D is made up of soft labeling, soft-hard labeling, and hard labeling in perception and cognition degrees, respectively.
Unsupervised perception and supervised perception could be explained by Gestalt principles (as is illustrated in section 2.2), which are fallen into soft labeling.
Cognition-driven labeling in C is categorized into hard labeling, which is frequently adopted in applications of goal orientations.
Furthermore, soft-hard labeling lies between soft and hard labeling, which utilizes both perception and cognition in labeling tasks.
The interaction-label idea, embedded in D, is widely used in various technologies and applications.
The subsequent work of this paper is focus on soft labeling.

\subsection{Perceptual Visual Interactive Learning (PVIL) Framework}
\begin{figure}[htbp]
	\centerline{\includegraphics[width=2.3in]{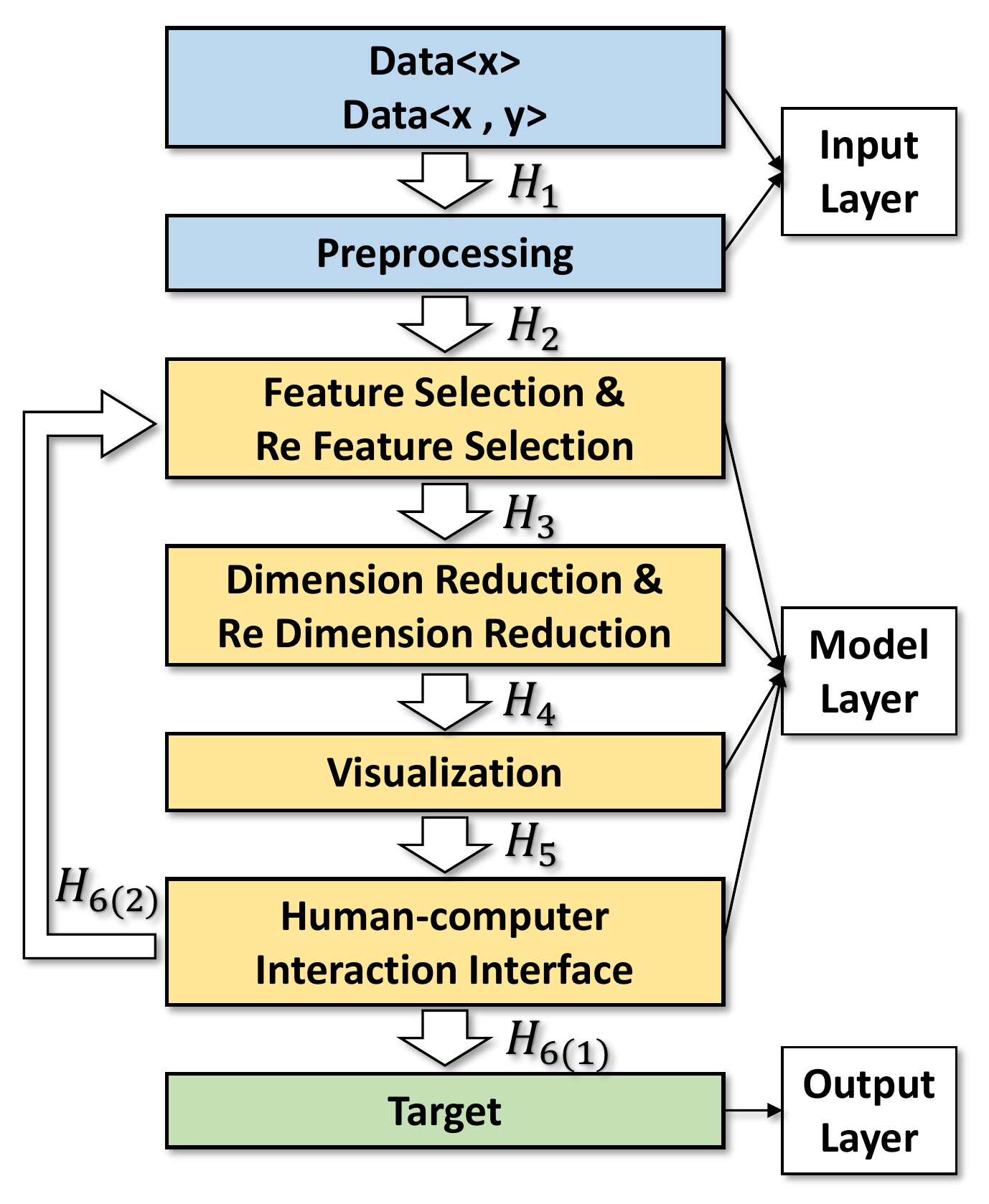}}
	\caption{Perceptual visual interactive learning framework}
	\label{fig5}
\end{figure}

An important purpose and application of interactive learning is classification.
This paper proposes a perceptual visual interaction framework for classification problems (Fig. \ref{fig5}). PVIL, a perception based VIL method, is to solve the problem that high-dimensional space cannot be perceived and labeled by human vision in SLQ. The framework achieved high accuracy in the case of SLQ classification. PVIL is introduced as follows.

PVIL is divided into three levels: the input layers, the model layers, and the output layer.
The input layers include data initialization and preprocessing, which is detailed by equation (\ref{eq1}) and (\ref{eq2}) in subsection 2.1.
Labeled data and unlabeled data are initialized by Eq. (\ref{eq8}) as follows.
\begin{equation}
\label{eq8}
H_1 =
\left[
\begin{array}{ccc}
X_m \\
X_n
\end{array}
\right]
label =
\left[
\begin{array}{ccc}
0 \\
Y_n
\end{array}
\right]
\end{equation}

And preprocessing is denoted as Eq. (\ref{eq9}).
\begin{equation}
\label{eq9}
H_2 = Pro(H_1)
\end{equation}
where Pro = \{Normalize, Regulation,\dots\} In general,
$H_1, H_2 \subseteq X_{(m+n)*h}$.

The samples here include scatter, image, music, video, etc.
In this paper, we adopt scatter and image as examples (part1, part2 is scatter, part3 is image) and the pre-processing part performs appropriate scale transformation for different sample types.
For example, the image is normalized \cite{Irizarry2003Exploration} to a standard normal distribution.
The model layers contain feature selection, dimensionality reduction, visualization and human-computer interaction.
Feature selection/extraction (Eq. (\ref{eq10})) refers to the representation of original data\cite{Guyon2003An} , such as the scale invariant feature transform (SIFT) feature \cite{Ng2003SIFT}, the histogram of oriented gradient (HOG) feature \cite{Dalal2005Histograms}, and the deep convolution feature \cite{Romero2016Unsupervised} in image understanding.
\begin{equation}
\label{eq10}
H_3 = FS(H_2)
\end{equation}
where $H_2\subseteq X_{s*h}$ then $H_3\subseteq X_{s*t_1}$.
As for dimensionality reduction (Eq. (\ref{eq11})), it refers to reducing the data dimensions on the premise of maintaining sample discriminative information as much as possible in classification issue.
\begin{equation}
\label{eq11}
H_4 = Dr(H_3)
\end{equation}
where $H_4 \subseteq X_{s*t_2}$.
Dimensionality reduction methods can fall into many categories,
such as linear methods principal component analysis (PCA) \cite{Mackiewicz2004Principal}, maximal similarity embedding (MSE) \cite{Feng2013Maximal}, and nonlinear ones  isometric mapping (ISOMAP) \cite{Balasubramanian2002The}, Locally Linear Embedding (LLE) \cite{Roweis2000Nonlinear}, local tangent space alignment (LTSA) \cite{Zhang2004Principal}, t-distributed stochastic neighbor embedding (T-SNE) \cite{Maaten2008Visualizing}, symmetric positive definite (SPD) manifold dimension reduction \cite{Harandi2017Dimensionality}, path based Isomap \cite{Najafi2016Nonlinear} and so on.
In our proposed PVIL, visualization (Eq. (\ref{eq12})) aims to reducing high-dimensional data to dimensions that humans can perceive, typically two-dimensional or three-dimensional.
\begin{equation}
\label{eq12}
H_5=Vis(H_4)
\end{equation}

Generally speaking, visualization can be treated as a special application of dimensionality reduction, that is $Vis=DR_2,$ $H_5 \subseteq X_{s*2}$.
In addition, the human-machine interface is used to control the loop of the above process (Fig. \ref{fig5}), which interact with user directly.
It judges whether multiple dimensional reduction and feature selection are required based on user input.
The algorithm processing steps of the human-computer interface are listed in Algorithm. \ref{alg:Framwork1}.

\begin{algorithm}[]
	\caption{Human-Computer Interface}
	\label{alg:Framwork1}
	\begin{algorithmic}[1]
		\REQUIRE ~~\\
		$H_5 \subseteq X_{s*2}$
		
		\ENSURE ~~\\
		$LabelOrd[LabelIndex_{k*1}, Label_{k*1}]$
		\STATE {$res = LabelIndex_{k+1} \cap X_n$}
		\IF {$Label[res].argmax().count()<Label[res].count * \eta $}
		\item {$H_2 = Pro(H_1[res, :])$}
		\item {JUMP TO $H2$}
		\ELSE
		\item {$LabelOrd[LabelIndex_{k*1},1]=Label[res].argmax()$}
		\ENDIF
		\IF {not $Smit$}
		\item {JUMP TO $H5$}
		\ENDIF
		\STATE \textbf{END}
\end{algorithmic}
\end{algorithm}

where $H_5$ is the result of visualization. We illustrate that $LabelIndex_{k*1} \subseteq \{1, 2, 3,\dots ,k\}$, $Label_{k*1}$$\subseteq$$\{c_1, c_2, \dots ,c_k \}$, $Smit\in$ $\{0, 1\}$, where $Smit$ indicates whether the user completed the experiment, $\eta\in [0.8, 0.9]$ is a hyper-parameter.
At last, the output layer saves class information calculated in the previous layer.

\subsection{MDR Visualization using PVIL}

\begin{figure}[htbp]
	\centerline{\includegraphics[width=3.3in]{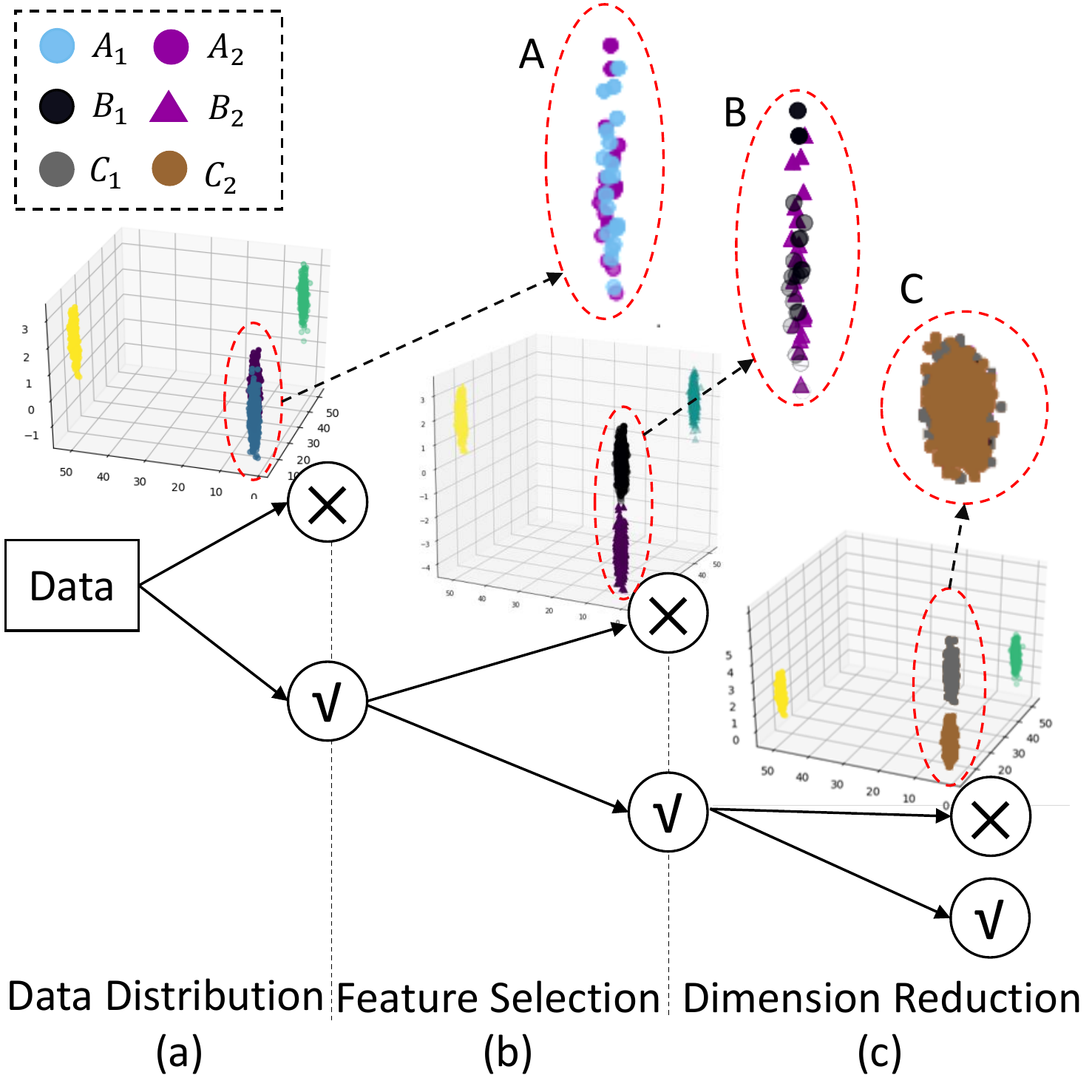}}
	\caption{Three cases of data overlap in low-dimensional space}
	\label{fig3}
\end{figure}

Dimensionality reduction which preserves the important relationship between samples is a key approach of implementing PVIL for completing interactive tasks with human perception.
However, visualization utilizing dimensionality reduction may lead to some incorrect results.
For example, data overlapping makes human perception inefficient in low-dimensional space (Fig. \ref{fig3}).
In the case of (a) in Fig. \ref{fig3}, Data Distribution: data overlaps itself in original space (see class A1 and class A2 in Fig. \ref{fig3}). Any feature is difficult to separate it. For case (b) in Fig. \ref{fig3}, Feature Selection: the origin dataset is separable in high-dimensional space, or original data (e.g. text or image separable). We need to represent the dataset by a feature extraction\footnote{Here, feature extraction means computing a feature by domain knowledge. For example, feature extraction of an image uses perceptual uniform descriptor (PUD) feature\cite{Liu2017Perceptual}.} method. However, it is difficult to separate overlapping samples by using the inappropriate feature for human interaction (e.g. class B1 and class B2 in ellipse B  of Fig. \ref{fig3}). (c) Dimension Reduction: If the samples are separable in the first two stage (a) and (b) of Fig. \ref{fig3}, inappropriate dimensionality reduction method may also leads to data overlapping (see class C1 and class C2 in ellipse C of Fig. \ref{fig3})\cite{Liu2015Scatter}.

\begin{figure}[htbp]
	\centerline{\includegraphics[width=3.3in]{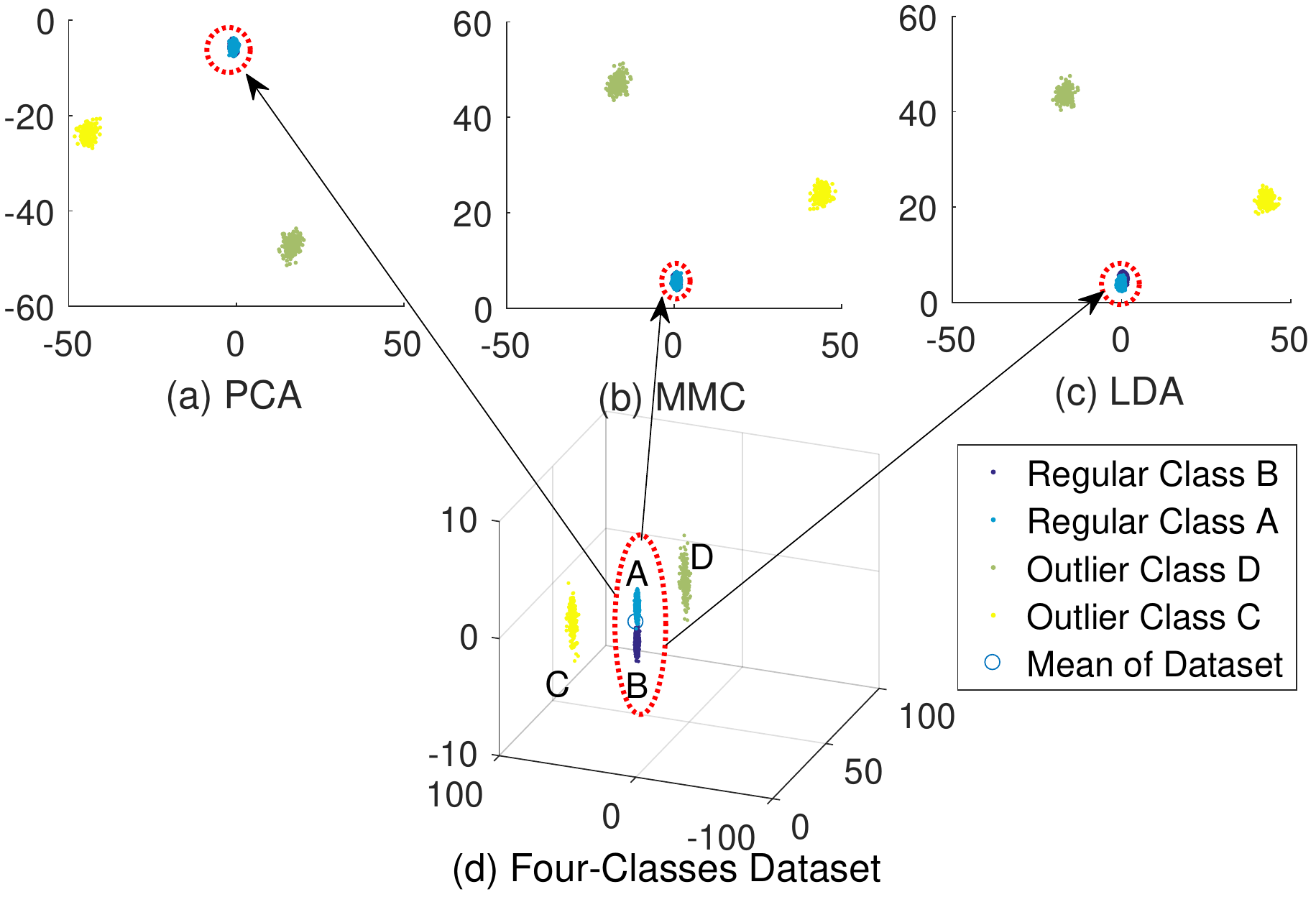}}
	\caption{The four classes dataset}
	\label{fig_fourclass}
\end{figure}

\begin{figure*}[htbp]
	\centerline{\includegraphics[width=6.5in]{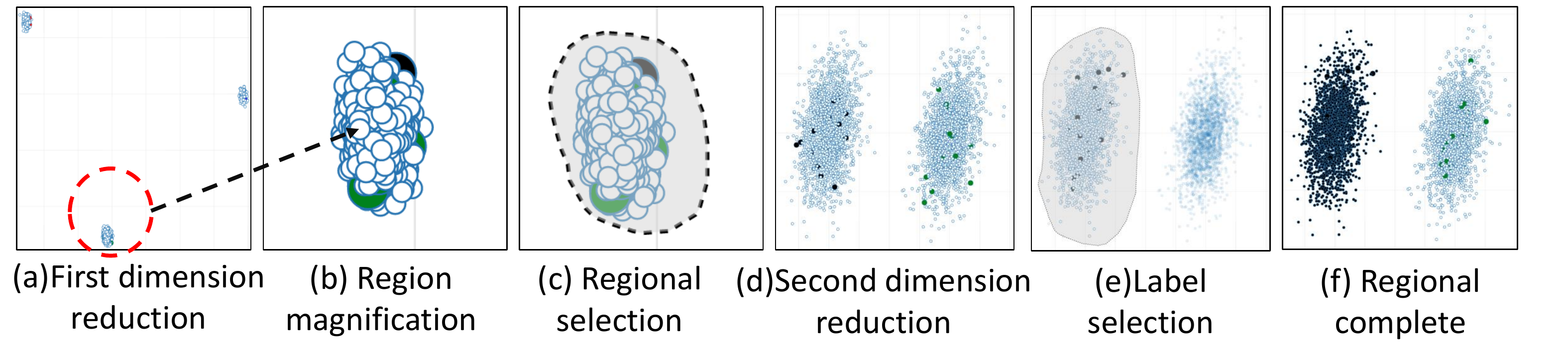}}
	\caption{The process of the MDR}
	\label{fig4}
\end{figure*}

\subsubsection{Compatibility problem of dimension reduction visualization and data distribution }

In this subsection, we will illustrate why dimension reduction and data distribution combining lead an inappropriate projection subspace in PVIL (see the case of Fig. \ref{fig3}). We point out that model structure of dimension reduction method should fit the data distribution. Here, we use PCA as a dimension reduction method, the optimization problem of PCA can be written by Eq. (\ref{eq111}) as follows.
\begin{equation}
\label{eq111} W_{opt}^{PCA} =\mathop {\arg \max }\limits_W
tr(W S_t W^T)
\end{equation}

where $W \in \mathds{R}^{D\times d}$, $W^TW=I$, and $S_t=\hat{X}\hat{X}^T$ denotes covariance matrix of $X$, where $\hat{X}=1/\sqrt{n-1}[x_1-\mu,\cdots ,x_n-\mu]$, $\mu=1/n\sum_{i=1}^{n}x_i$.
$W_{opt}^{PCA} $ can be got by the eigenvectors corresponding the first $d$ largest eigenvalues of $S_t$. PCA aims to project the original (or after feature selection) dataset to a low-dimensional subspace and preserves the largest variance of samples, which makes data overlapping by the incompatibility between model and data distribution in four classes dataset or similar conditions (see Fig. \ref{fig_fourclass}, and details can be refer to subsection 4.3 or reference\cite{Liu2015Scatter}). This problem also happens in the scene of LDA and MMC\cite{Liu2015Scatter} (see Fig. 5 (b) and (c)). Therefore, we can conclude the problem that \emph{compatibility between dimension reduction and data distribution/feature selection is important for PVIL}.

\subsubsection{The process of Multi-dimensionality Reduction Visualization}

In order to solve the problem of section 3.3.1, this paper proposes a ``multiple feature selection and multiple dimensionality reduction\footnote{We only use multi-dimensionality reduction visualization because multiple feature selection utilizes the similar approach and get the similar results in the rest of this paper.}'' method (Fig. \ref{fig4}), which adopts two or even multiple interactions to compensate for the shortcomings of once feature selection or dimensionality reduction. The process of the proposed multi-dimensionality reduction visualization can be listed as follows.

(1)The user can get the visual results after initialization and dimensionality reduction (see Fig. \ref{fig4} (b)). Meanwhile, the overlapping low-dimensional data of the features will be perceived by the user and transmitted to the interaction framework (See Fig. \ref{fig4} (c)).

(2)The interaction framework will generate a new subspace for low-dimensional data visualization of original feature space (See Fig. \ref{fig4} (d)). The user can decide whether current visualization results are reasonable or not.Repeat this process of perception until the final subspace and appropriate features are selected.

Regarding dimensions, the methods are similar to features. But dimensionality reduction also involves the imbalance of data distribution.
Multiple dimensionality reduction only selects overlapping area(s) of visualization graph (e.g. Fig. \ref{fig3} or Fig. \ref{fig4} ), so that the real subspace (e.g. Fig. \ref{fig4} (d)) which can achieve high performance in classification is not affected by the distribution of data.The detail of Compatibility problem of dimension reduction visualization and data distribution can be referred section 3.3.1.
In essence, the above method achieves the goal that linearly separable data in high-dimensional space is still holding in low-dimensional one. This is also the key idea of the perceptual visual interactive framework proposed in this paper.

\section{Experimental Results and Analysis}

To validate the effectiveness and efficiency of VIL and the proposed PVIL framework, experiments are conducted on synthetic datasets and real-world dataset. Note that, user participation are required in the the process of VIL. Details of the experimental settings are introduced in section \ref{expintro}. To evaluate the influences of PVIL on labeling, experiment is conducted on a synthetic dataset and label propagation (LP) is adopted as the comparing method (Details in section \ref{exp1}). Moreover, dimensional reduction experiment is also conducted to fully reveal the effectiveness of MDR (Details in section \ref{exp2}). The proposed PVIL aims at combining computer's sensitivity of detailed features and human's overall understanding of a task. Hence, classification experiment is conducted to evaluate the effectiveness of PVIL (Details in section \ref{exp3}).

\subsection{Experimental Introduction}

\label{expintro}
\begin{figure}[htbp]
	\centerline{\includegraphics[width=3.4in]{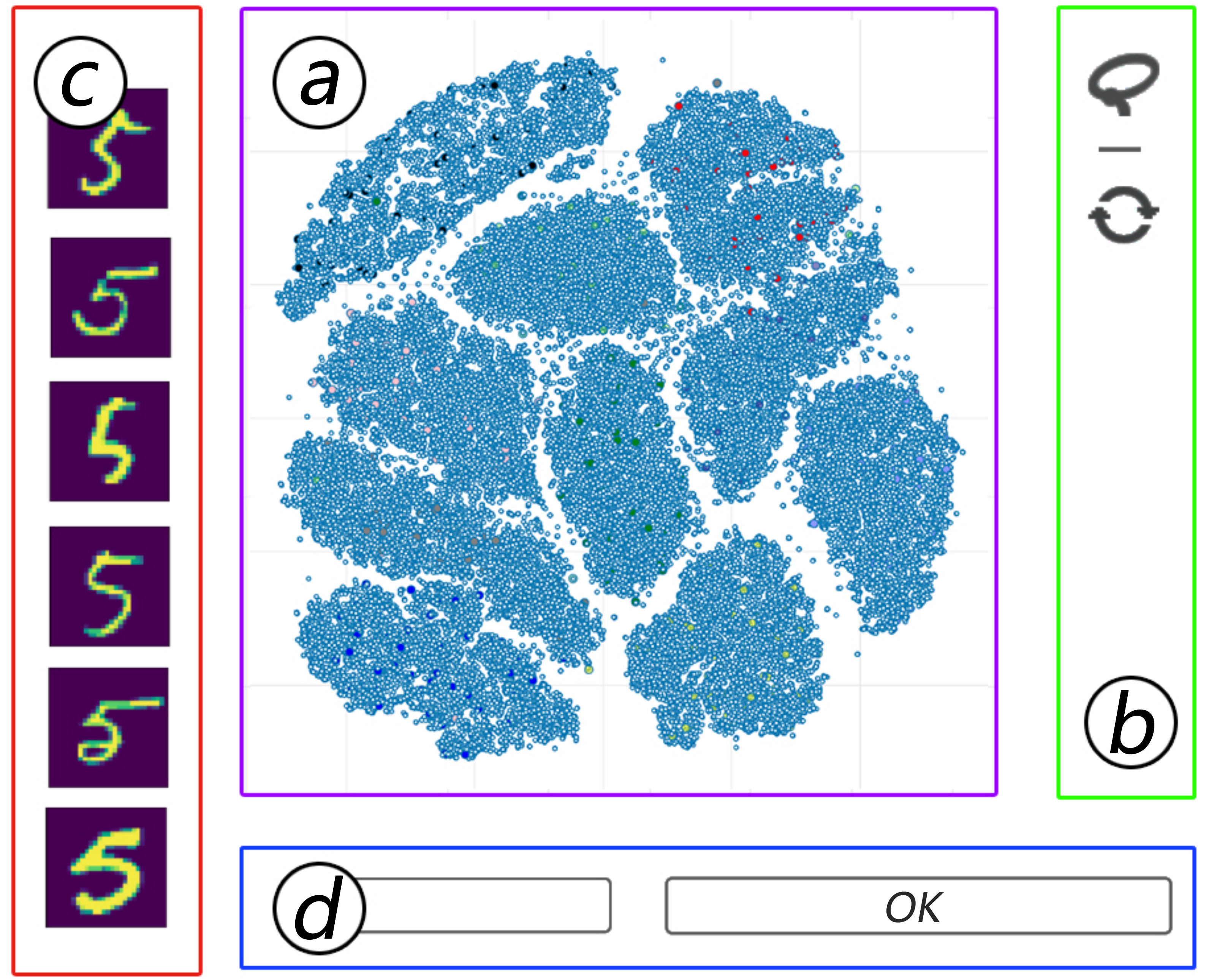}}
	\caption{Interactive interface of PVIL. The use of the interface is divided into the following four parts: (a) Observing the visual result. (b) Using the circle selection tool to perform the element principle. (c) Viewing the label value in the selected element. (d) Confirm this selection.}
	\label{fig6}
\end{figure}

In our experiment, 20 participants (12 males, 8 females) are recruited whose ages are ranging from 18 to 23.
Each participant has a normal vision ability and receives college education over one year or more. In our experiment, participants are required to have no idea of the process of VIL.
Even some of the participants are interested in interactive learning, no one has expertise in the field, or have seen the detailed distribution of the dataset.
An experimental tool  is designed and arranged on a high-performance server in our experiment to realize user-machine interface. Each participant connects to the tool through a web page.
The use of the experimental tool can be  divided into 4 steps, as shown in Fig. \ref{fig6}

To ensure the completeness and objectivity of the experiments, experimental process is detailed introduced to the participants.
In particular, the best labeling results will be rewarded.
During the experiment, each user is required to record the labeling strategy used during the labeling process, such as how to handle critical data, choose aggressive or conservative ways to complete the data labeling task. The strategies are recorded by the participants on an issued form. The entire experiment time is approximately 90 minutes.
We mainly select the following two indicators to judge the differences between different labeling methods:
(1) Unlabeled rate is defined as the percentage of the number of unlabeled data ($DATA_{ublabeled}$) to the number of  total data ($DATA$), which is used to show the proportion of the labels. It is denoted as $R_{unlabeled} = DATA_{unlabeled}/{DATA}$.

(2) The data accuracy is measured by the F1-measurement, which is commonly used in  classification experiments.
Given a set of data, according to the classification criteria, we count number $N_{TP}$ of True Positives, False Positives ($N_{FP}$) and False Negatives ($N_{FN}$).
We compute the following performance measure: recall ($R = N_{TP}/(N_{TP}+N_{FN})$), precision ($P = N_{TP}/(N_{TP}+N_{FP})$) and $F_1$ score ($Acc_{f1}=2PR/(P+R)$).

\subsection{Labeling Performance Comparison between LP and PVIL on Synthetic Dataset}

\label{exp1}

We conduct an experiment on a two moons synthetic dataset\footnote{``$x$'' shape is only used for illustrating the good continuity of human perception.}(as is shown in Fig. \ref{fig8} (a)) to assess the labeling performance of LP and PVIL.
The origin dataset is consisted of 200 samples with two classes\cite{Zhou2003Learning}. We also increase the amount of samples up to $N(10^5)$, where $N(\cdot)$ indicates samples' amount.

\begin{figure}[htbp]
	\centerline{\includegraphics[width=3.4in]{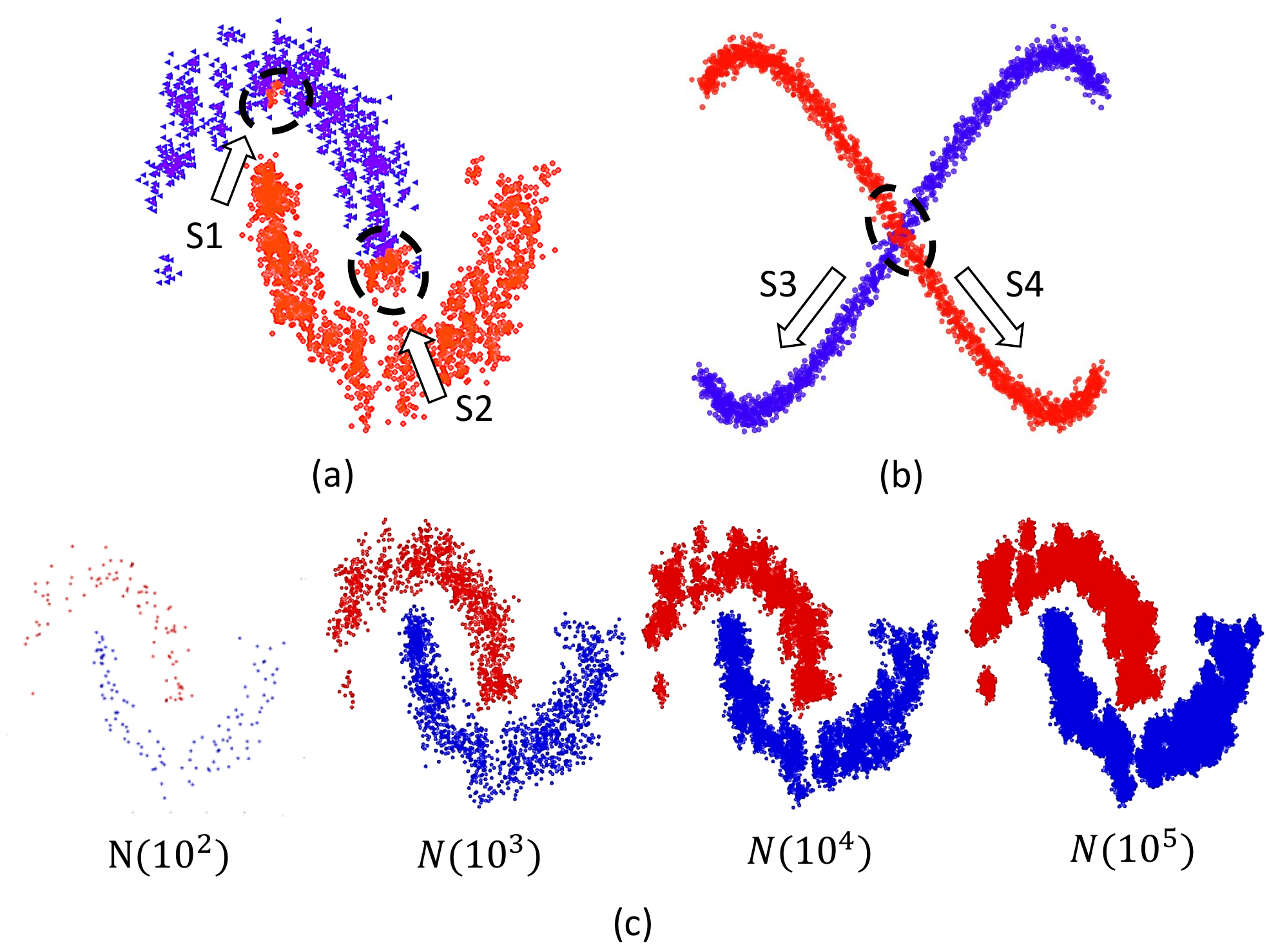}}
	\caption{Two shapes datasets including two moons and ``$x$'' shape. Label propagation may select wrong direction of propagation, as shown in (a) (b), and (c) shows the shape of the manifold at different levels of data. The amount of samples is $10^2$, $10^3$, $10^4$ and $10^5$ respectively.}
	\label{fig7}
\end{figure}

\begin{figure}[htbp]
	\centerline{\includegraphics[width=3.5in]{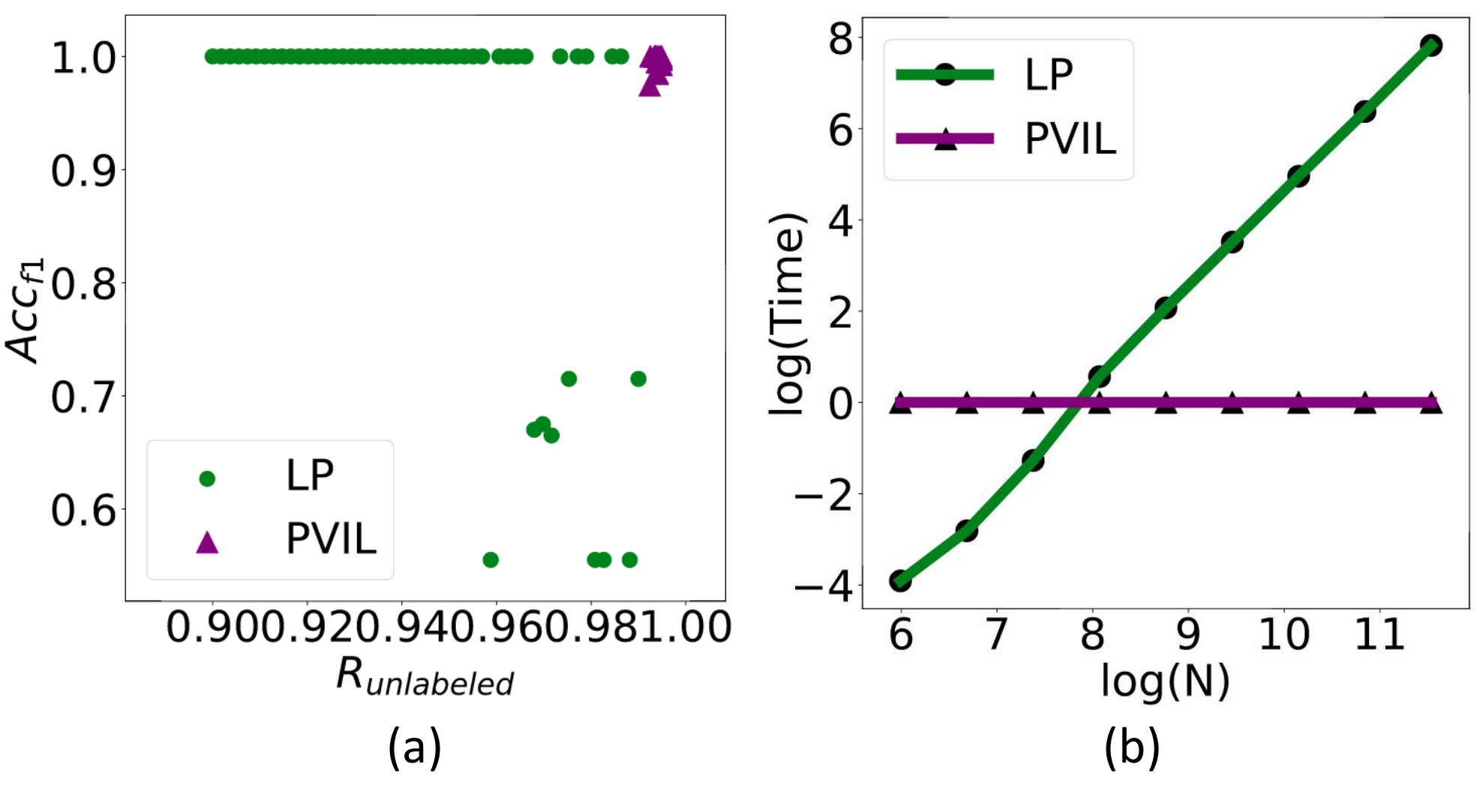}}
	\caption{The comparison of LP and PVIL in accuracy and time consuming. (a)The accuracy of LP and PVIL at different unlabeled quantities. (b)The time consuming of LP and PVIL with different data levels.}
	\label{fig8}
\end{figure}

The performance of LP algorithm highly relies on adequate labels.
For LP algorithm, the similarity calculation for samples is based on ranking method.
During the process of ranking, to label a new unlabeled sample depends on the existing labeled samples (training set).
LP algorithm, thus, works well when enough labels exist.
As is shown in Fig. \ref{fig8}(a), when the unlabeled rate $R_{unlabeled}$ is below $95\%$, the $Acc_{f1}$ is near $100\%$.
However, when the amount of labels is inadequate, the performance of LP is unsatisfactory.
For lack of the neighbor information, it is difficult to find a suitable neighbor size ($k$) in the process of LP.
Furthermore, the process of propagation would be misled easily when neighbor information is inadequate.
In the case of insufficient labels, it is possible for the LP to select a propagation direction similar to S1, S2 (as shown in Fig. \ref{fig7} (a)).
Worse still, since the choice of improper propagation direction, wrong labeling results would be calculated.
It can be seen in Fig. \ref{fig8} (a), when $R_{unlabeled}$ is above $95\%$, the $Acc_{f1}$ is under $60\%$ in certain processes of LP.
The similar situation of Fig. \ref{fig7} (a) is particularly evident in the data manifold of Fig. \ref{fig7} (b).
As the distance of two manifolds is relative small in the directions of $S3$ and $S4$ (as shown in Fig. \ref{fig7} (b)), it is difficult to select a proper subspace in the process of LP algorithm.

In contrast, PVIL has a distinct advantage over LP, which derives mainly from the human visual perception ability.
During the process of similarity calculation, closed or warped data can be inferred as an uniform manifold by PVIL.
Euclidean distance, thus, will be put on the back burner in the decision process.
As is shown in Fig. \ref{fig7} (a), improper propagation directions (S1 and S2) are ignored by PVIL.
Similarly, proper propagation directions of S3 and S4 in Fig. \ref{fig7} (b) are  chosen, respectively.
As illustrated in Fig. \ref{fig8} (a), the PVIL algorithm achieves the $Acc_{f1}$ of $99\%$ under $R_{unlabeled}$ around $99\%$.
Hence, the PVIL could outperform LP on dense distribution and sparse classes dataset.

\begin{table}[htbp]\centering
	\caption{Time cost comparison between LP and PVIL}
	\setlength{\tabcolsep}{10mm}
	\label{table1}
\begin{tabular}{ccc}
	\toprule
	& LP& PVIL\\
	\midrule
	$N(10^2)$& $10^{-3}s$& $<10^1s$\\
	$N(10^3)$& $10^{-1}s$& $<10^1s$\\
	$N(10^4)$& $10^1s$& $<10^1s$\\
	$N(10^5)$& $10^3s$& $<10^1s$\\
	$N(10^6)$& $None$& $<10^1s$\\
	\bottomrule
\end{tabular}
\end{table}

PVIL realizes labeling by utilizing human perception, which is quite time-efficient regardless of the amount of samples.
Hence, PVIL will be more time-saving than LP under a large-size dataset.
The time complexity of LP is $O(n^2)$, while the time complexity of PVIL is $O(1)$.\footnote{For convenience, the time complexity of dimensionality reduction is ignored in this experiment.}
With the increases of the data size in the logarithmic axis, the time cost of LP increases linearly (as shown in Fig. \ref{fig8}(b)).
On the other hand, the time cost of PVIL remains basically the same (as shown in Fig. \ref{fig8}(b)).
Quantized comparison of time cost is shown in Table. \ref{table1}.
When the amount of samples is below $10^4$, both the LP and the PVIL can complete the task less than 10s.
When the amount of samples reaches $10^5$, the time cost of LP is over 100 times than that of PVIL.
When the amount of samples reaches $10^6$, the space complexity of LP will also reach an unacceptable level apart from the time complexity. The problem limits the application of LP on large scale dataset.
By contrast, PVIL could work normally regardless of the amount of samples.

\subsection{Labeling Performance of MDR}

\label{exp2}

\begin{figure}[htbp]
	\centerline{\includegraphics[width=3.4in]{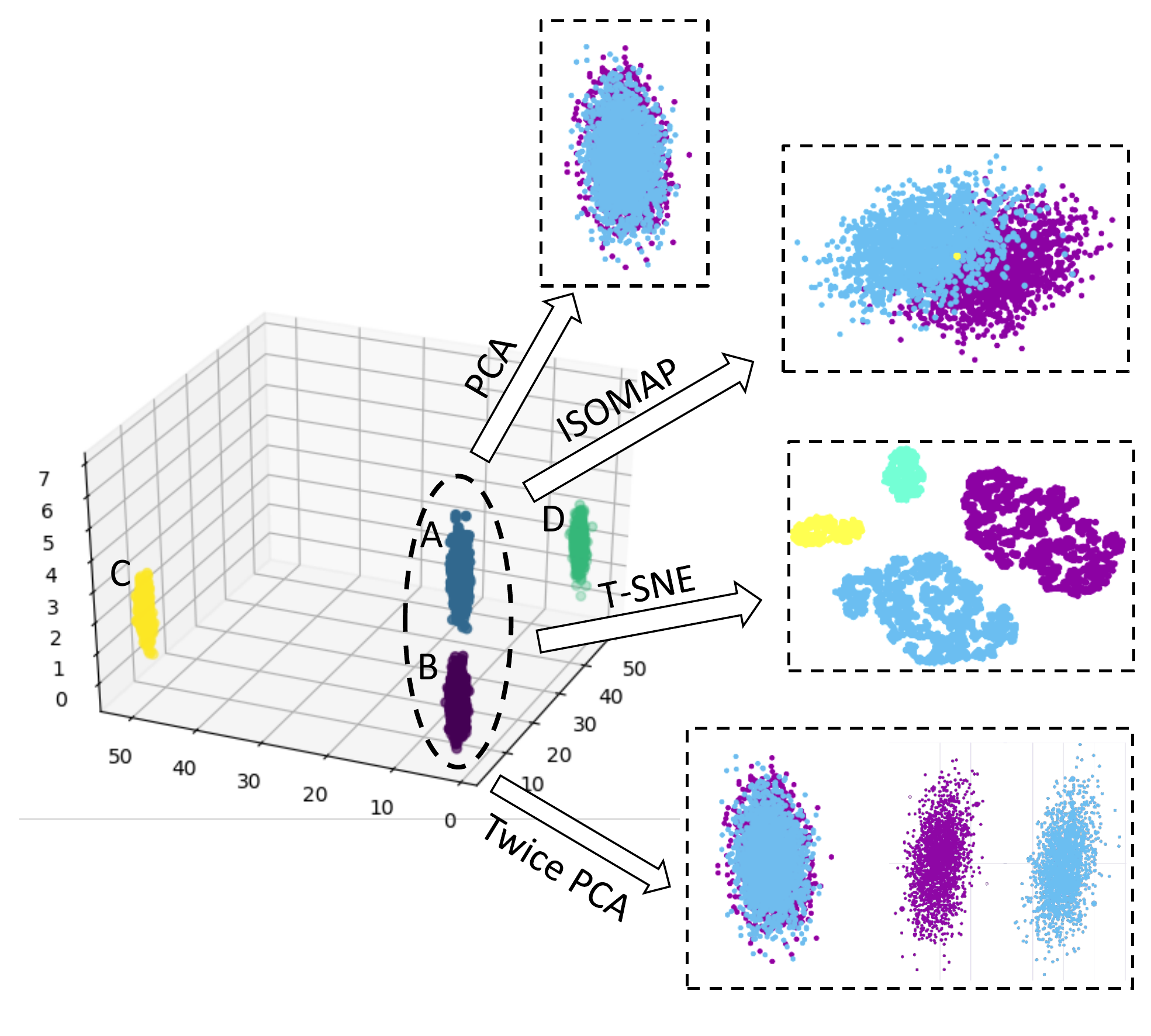}}
	\caption{The dimension reduction (PCA, ISOMAP, T-SNE and Twice PCA) results of four-classes dataset. The dataset cannot be separated by one-time global methods (PCA and ISOMAP).}
	\label{fig9}
\end{figure}

To review the performance of MDR with PVIL, we compare it with other unsupervised dimensional reduction methods (PCA, ISOMAP and T-SNE).\footnote{The task is difficult for supervised dimensionality reduction methods such as linear discriminant analysis (LDA) \cite{izenman2013linear} due to the fact that the unlabeled rate $R_{unlabeled}$ of  the dataset (99.5\%) is too high.}
A four-class synthetic dataset containing 4800 instances (as is shown in Fig. \ref{fig9}) is involved in this experiment.
The covariance matrix and the means of the four classes dataset are listed in Table. \ref{table_para}\cite{Liu2015Scatter}.
And the scatters of the dataset are showed in Fig. \ref{fig_fourclass} (d).

\begin{table}[htbp]

\begin{center}
\caption{The parameter settings of four classes dataset}
\setlength{\tabcolsep}{6mm}
\begin{tabular}{cc}
\toprule
Mean& Covariance Matrix  \\
\midrule
\raisebox{-3.00ex}[0cm][0cm]{[5,3,1][5,3,5][50,3,2][5,50,2]}&
\raisebox{-3.00ex}[0cm][0cm]{$\left[ {{\begin{array}{*{20}c}
 {0.3} \hfill & {0.04} \hfill & {0.06} \hfill\\
 {0.04} \hfill & {0.2} \hfill & {0.05} \hfill\\
 {0.06} \hfill & {0.05} \hfill & {0.2} \hfill\\
\end{array} }} \right]$} \\
&\\
&\\
\bottomrule
\end{tabular}
\label{table_para}
\end{center}
\end{table}

\begin{table}[htbp]\centering
	\caption{Comparison of accuracy and time complexity of PCA, ISOMAP, T-SNE and MDR(Twice PCA)}
	\setlength{\tabcolsep}{8mm}
	\label{table2}
	\begin{tabular}{ccc}
		\toprule
		& $Acc_{f1}$& $Time(s)$\\
		\midrule
		PCA& 0.583& 0.004\\
		ISOMAP& 0.812& 6.204\\
		T-SNE& 0.999& 55.536\\
		MDR(Twice PCA)& 0.994& 0.010\\
		\bottomrule
	\end{tabular}
\end{table}

As detailedly illustrated in subsection 3.3, overlap may exist in the dimensionality reduction result of some datasets.
The primary reason for the overlap is the existence of outliers.
These outliers mislead the process of dimensionality reduction by affecting the choice of subspace.
It can be seen in Table. \ref{table2} that the $Acc_{f1}$ of the PCA and ISOMAP are $58\%$ and $81\%$, respectively.
The result indicates that the data in our four-classes dataset can not be separated by one-time global methods (PCA and ISOMAP), and the dimensional reduction process is influenced by outliers (As shown in Fig. \ref{fig9}).

The MDR method could capture semantic information better than that of one-time global methods.
Concretely speaking, the distribution of outliers can be located by the first order dimensionality reduction with the help of VIL.
Furthermore, the existence of the outliers can be ignored during the process of the second order dimensionality reduction.
The advantage of MDR mainly derives from the localization of outliers.
As is shown in Table. \ref{table2}, the $Acc_{f1}$ of MDR method is around $99\%$.
MDR achieves comparable performance with T-SNE, which is much better than the other one-time global methods (PCA and ISOMAP).

The time complexities of these dimensionality reduction algorithms are discussed as follows.
Note that, the time complexity of MDR(Twice PCA) is near PCA, which can be regarded as PCA algorithm.
To realize dimensionality reduction for $n$  samples and $p$ features, the time complexity of PCA is $O(p^2n+p^3)$.
The optimal time complexity of ISOMAP algorithm is $O(n^3)$ \cite{Balasubramanian2002The}.
As for T-SNE, the solution must be approximated through gradient descend iterations, which is quite time consuming.
In our experiment, the time consumption of the three methods is $Time_{PCA} < Time_{ISOMAP} < Time_{T-SNE}$.
In general, only the PCA algorithm can meet the time requirement of interactive learning in practice.

\subsection{Performance of PVIL on Real-World Dataset}

\label{exp3}

In this experiment a real-world handwritten dataset (MNIST) is adopted  to evaluate the effectiveness of PVIL.
The MNIST dataset is made up of 60,000 training instances and 10,000 test instances. Each instance is a 0 to 9 handwritten digit image, which is represented by a  $28\times28$ matrix. In our experiment, the matrix is transformed to a 784-dimensional vector to represent the original grayscaling image.

\subsubsection{Labeling Performance Comparisn between LP and PVIL on Real-World Dataset}
To evaluate the labeling performance of LP and PVIL, the unlabeled rate is set as $R_{unlabeled}$ of $99.5\%$ in our experiment. In the experiment, the labeling results of 20 participants using PVIL are recorded.
For fair comparison, the LP algorithm is executed 20 times with random initialization, the result of which in each time is also recorded.

\begin{table}[htbp]\centering
	\caption{Comparison of the accuracy between LP and PVIL}
	\label{table3}
	\begin{tabular}{ccccc}
		\toprule
		& $R_{unlabeled}$& $Acc_{f1}(worst)$ & $Acc_{f1}(average)$&$Time(s)$ \\
		\midrule
		LP& $99.5\%$&  0.0995 & 0.0999& 1776.1\\
		PVIL& $99.5\%$ &  0.9504 & 0.9654& 494.7\\
		
		\bottomrule
	\end{tabular}
\end{table}

The semantic gap is one of the hard-to-solve problems in  computer vision.
It has been explained in subsection 2.2 that the perception ability of computers is weaker than that of humans for closed or warped data in visualization.
The advantage of human perception is particularly evident in the labeling tasks of real-world dataset.
As is shown in Table. \ref{table3}, the labeling accuracy of LP for 99.5\% $R_{unlabeled}$ data is about 10\%, which is far from the request of labeling tasks.
While PVIL method can better tolerate overlapping and bending information through interaction, which achieves an accuracy of 96.54\% on average for the 99.5\% $R_{unlabeled}$ data.


In terms of computational efficiency, the measurement of relationships between samples is isolated with the LP algorithm.
To obtain the weight map, the measurement of the similarity between every two samples is calculated by a kernel function, which results in high computation cost.
As for the MNIST dataset, LP  takes an average of 1776.1s in time consumption, which is unacceptable in most real-world applications.
Note that, the human perception for a large number of samples is holistic. The process of human perception can skip the specific neighborhood similarity calculations, and greatly reduce the time of constructing a weight map.
However, it takes some time for user to interactive with computers.
In our experiments, 20 participants spent an average of 494.1s to complete a group of labeling.
It is obvious that PVIL is superior to LP in efficiency, which ensures the real-world applications of PVIL.

\subsubsection{Classification Performance Comparison between PVIL and Ground-truth Labels}

\begin{figure}[htbp]
	\centerline{\includegraphics[width=2.4in]{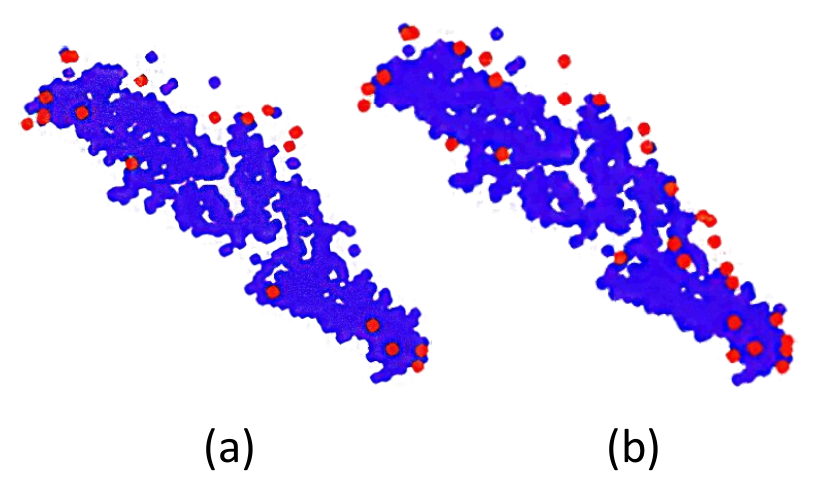}}
	\caption{Classification Performance of ground-truth labels (a) and PVIL labels (b). Blue points are True Positives and red ones are False Negatives.}
	\label{fig10}
\end{figure}

To further evaluate the performance of the labeling results of PVIL in real-world application, a classification experiment is conducted.
Several mainstream classifiers, including  logistic regression (LR)\cite{harrell2015ordinal}, gradient boosted decision trees (GBDT)\cite{friedman2001greedy}, random forest (RF)\cite{liaw2002classification} and convolutional neural network (CNN)\cite{lecun1998gradient}, are adopted in our experiment. \footnote{LeNet is chosen as the convolutional network model rather than deeper one such as VGG16\cite{Simonyan2014Very} or ResNet\cite{He2016Deep} for the reason that the the task complexity is relatively low.}
The classifiers are trained by the labels obtained by PVIL algorithm and ground-truth labels respectively.\footnote{It is difficult to continue the classification experiment by LP labels since the $Acc_{f1}$ of the LP labels in subsection 4.4.1 is extremely low ($<$10\%).}


\begin{table}[htbp]\centering
	\caption{Comparison of the ground-truth label and the PVIL label in the test set after training by LR, GBDT, RF, and CNN.}
	\label{table4}
	\begin{tabular}{cccc}
		\toprule
		& Ground-Truth Labels& PVIL labels & Loss \\
		\midrule
		LR& 0.8600 & 0.8525& 0.075\\
		GBDT&  0.9390 & 0.9354& 0.0046  \\
		RF&  0.9602 & 0.9534&0.0068 \\
		CNN (LeNet)& 0.9901 & 0.9745 & 0.0156  \\
		\bottomrule
	\end{tabular}
\end{table}

The process of PVIL in acquiring labels can combine computer's sensitivity of detailed features and human's overall understanding ability of a task.
Hence, the labeling results of PVIL could reveal the actual labeling distribution better.
The experiment results (Table. \ref{table4}) show that PVIL labels achieve significant classification results on real-world dataset.
As is listed in Table. \ref{table4}, under the premise of $96.5\%$ label accuracy, the highest classification accuracy of PVIL could reach $97.5\%$ (CNN),
which is approaching the performance with ground-truth labels.
The advantage of PVIL mainly lies in the perception  of the backbone manifold, which can effectively ensure the high labeling accuracy.

As for the shortcomings, the error of PVIL mainly comes from the lack of accurate measurement ability.
Isolated samples, located in the boundary area of a manifold, is difficult to distinguish even for humans.
As is shown in Fig. \ref{fig10}, most errors of the PVIL labels are distributed in the boundary area.
However, It does not appear a significant loss of classification accuracy in our experiment for the reason that the isolated samples are relatively sparse.

\section{Conclusion}
The main purpose of this paper is to re-summarize the related technologies and applications of interactive learning from the perspective of VIL. We thus propose a framework called PVIL that structures an interaction strategy based on Gestalt principle. PVIL takes advantage of the idea of multi-dimensionality reduction to further boost the performance of visualization. The experiments validate that the PVIL framework significantly outperforms the labeling propagation for most visualization tasks. Furthermore, PVIL achieves promising classification results on real-world datasets.

In the future, we would attempt to develop a perceptual-cognitive framework with diversification for the aim of extending current PVIL to more learning scenes such as weak supervised learning and transfer learning.

\section*{Acknowledgements}
This study was funded by National Natural Science Foundation of People's Republic of China (61672130, 61602082, 91648205), the  National Key Scientific Instrument and Equipment Development Project (No. 61627808), the Development of Science and Technology of Guangdong Province Special Fund Project Grants (No. 2016B090910001). All the authors declare that they have no conflict of interest.

\ifCLASSOPTIONcaptionsoff
  \newpage
\fi



%
\bibliography{references}
\bibliographystyle{IEEEtran}

%




\end{document}